\documentclass{article}

\PassOptionsToPackage{numbers, compress}{natbib}

\usepackage[final]{neurips_2019}

\usepackage[utf8]{inputenc} 
\usepackage[T1]{fontenc}    
\usepackage{hyperref}       
\usepackage{url}            
\usepackage{booktabs}       
\usepackage{amsfonts}       
\usepackage{nicefrac}       
\usepackage{microtype}      
\usepackage{xcolor}
\usepackage{soul}
\usepackage{booktabs}
\usepackage{multirow}
\usepackage[linesnumbered,ruled]{algorithm2e}
\usepackage{subcaption}
\usepackage{scalerel,graphicx,xparse}

\title{Predicting Future Sales of Retail Products using Machine Learning}

\author{%
  Devendra Swami\\
  MS, Computer Science\\
  University of Southern California\\
  Los Angeles, CA 90007 \\
  \texttt{dswami@usc.edu} \\
   \And
  Alay Dilipbhai Shah\\
  MS, Computer Science\\
  University of Southern California\\
  Los Angeles, CA 90007 \\
  \texttt{alaydili@usc.edu} \\
   \And
  Subhrajeet K B Ray\\
  MS, Computer Science\\
  University of Southern California\\
  Los Angeles, CA 90007 \\
  \texttt{skray@usc.edu} \\
}

\begin{document}

\maketitle

\begin{abstract}

Techniques for making future predictions based upon the present and past data, has always been an area with direct application to various real life problems. We are discussing a similar problem in this paper. The problem statement is provided by Kaggle, which also serves as an ongoing competition on the Kaggle platform. In this project, we worked with a challenging time-series dataset consisting of daily sales data, kindly provided by one of the largest Russian software firms - 1C Company. The objective is to predict the total sales for every product and store in the next month given the past data.

In order to perform forecasting for next month, we have deployed eXtreme Gradient Boosting (XGBoost) and Long Short Term Memory (LSTM) based network architecture to perform learning task. Root mean squared error (RMSE) between the actual and predicted target values is used to evaluate the performance, and make comparisons between the deployed algorithms. It has been found that XGBoost fared better than LSTM over this dataset which can be attributed to its relatively higher sparsity. 
\end{abstract}

\section{Introduction}
Accurate sales forecasting is of utmost value to any organization, be it a small and medium enterprise or a Fortune 500 company. It provides an accurate estimation of company's top-line growth and can be utilized to prepare plans for near-term demand and supply actions. Also, it serves as a guidance to devise the bottom-line plans that keep up with the overall organizational goals for financial prudence. Finally, strategic decisions such as identification of key areas for short-term investment can also be made, taking cues from these forecasts. 

Interesting and difficult sales forecasting problems are pretty common. We are very fortunate that a similar problem statement titled "Predict Future Sales" is provided by Kaggle.~\cite{kaggle} As part of the problem statement, we need to work over a challenging time-series dataset consisting of daily sales data. A time series is a sequence of historical measurements of one or more observable variable(s) at equal time intervals. Time series can be studied for several purposes such as predicting future based on past knowledge, or understanding the latent variables behind the generation of measured values, or for simply providing a concise description of the salient features of the series. For this project, we need to forecast the total sales of every product and store combination for the next month, given the past data.   

We have started working on the problem by performing exploratory data analysis (EDA) over the provided dataset, in order to understand the dataset better. For this task, we took help from the already available notebooks on Kaggle. Soon, we found out that our dataset is a combination of time series and static data. Columns such as item name, item category, shop ID, etc. do not change with time whereas item count and item price do vary with time. Having a better idea of the training dataset, we started to look for ideal candidates to perform this learning task. We split our team into two groups to simultaneously cover two different objectives. One group started to read more kaggle notebooks in order to understand what has worked for most people. Another group searched for what can be some novel techniques, that can be applied on this problem.

With time, we have realized that applying XGBoost with lagged features did work out pretty well for many people in the past. Also, we have found that Autoregressive Integrated Moving Average (ARIMA) and LSTM based networks have provided quality results over similar forecasting tasks.~\cite{lstm} ~\cite{arima} More about this is discussed in the subsequent section on related work. To summarize, we have decided to try XGBoost, ARIMA, and LSTM as learning algorithms, and to use RMSE for evaluating their performance.




\section{Related work}

The big inspiration for our model selection process was drawn from an outstanding Kaggle Blog Post \textbf{\textit{"Time Series - ARIMA, DNN, XGBoost Comparison"}}.~\cite{kaggle_blog}

As the name says, the blog is about comparison of three models; ARIMA, DNN and XGBoost on Time Series Dataset. The format of the data set on which the comparisons were drawn largely matches the format of the dataset on which we had to make our predictions. And since all the above three models gave fairly appealing results in the blog, hence we were confident that we could not go wrong experimenting with these models if we did other parts of the job like data preparation and feature engineering right way.

The same blog also shows some good featuring engineering practices which suits best for each model. Hence, some of the ideas for feature engineering were also inspired from the same Kaggle Blog.

However, for some reason, we did not get impressive results using DNN with our first implementation. Meanwhile, we found another interesting paper which outlines how LSTM could be applied to a dataset of this format. More about the paper below, but the important thing to mention here is that instead of spending time playing around and fixing DNN, we thought to shift our focus on LSTM. Hence, our final set of models to experiment were; ARIMA, LSTM and XGBoost.

\begin{figure}[ht]
\vspace{-10 mm}
    \centering
  \includegraphics[width=1.0\columnwidth]{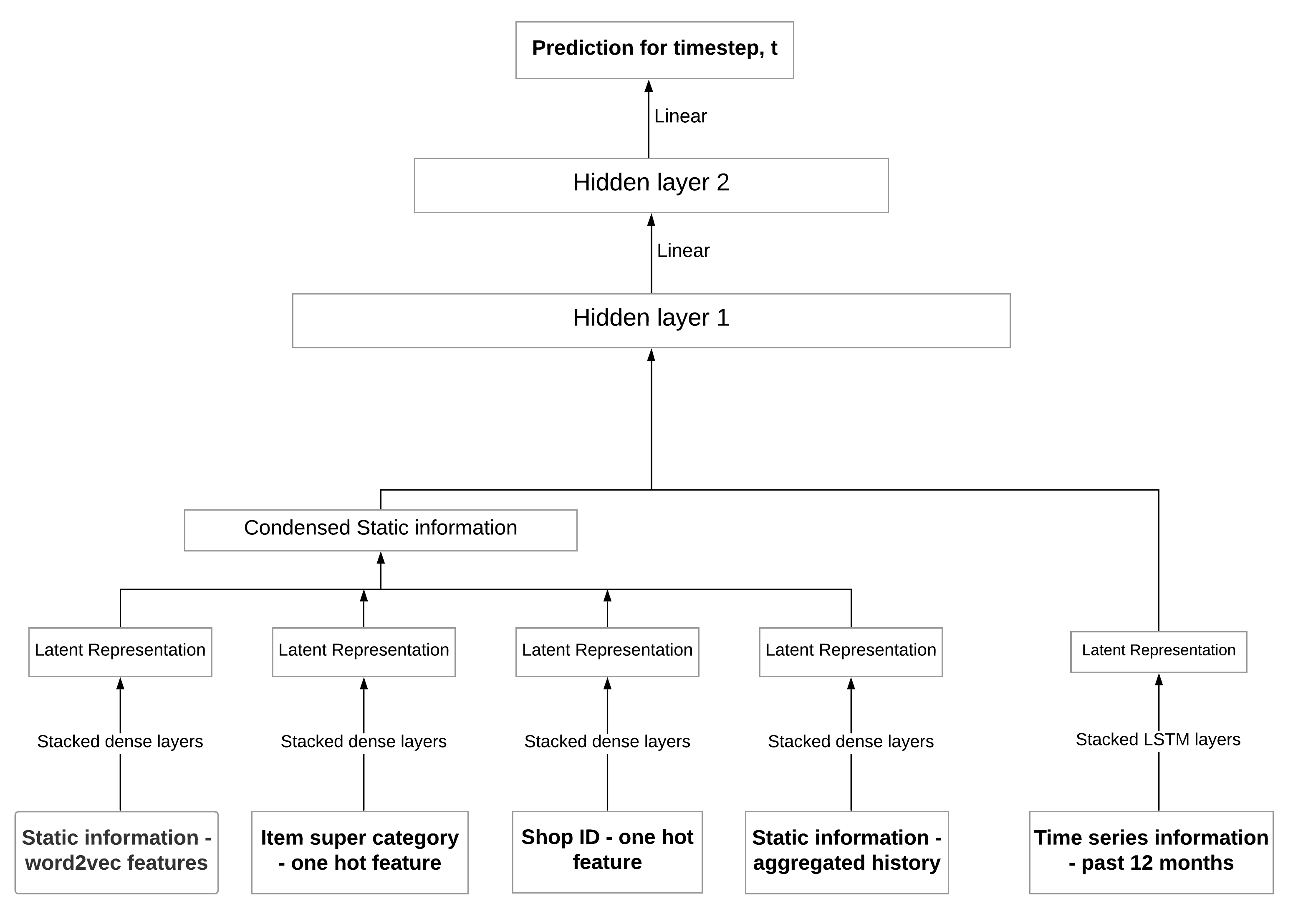}
  \caption{LSTM-based neural network architecture, combining static and dynamic features}
  \label{fig:lstm_arch}
\end{figure}

Esteban, et al. (2016) have demonstrated in their paper, how static and dynamic features can be efficiently modeled with a combination of LSTM and fully connected neural network architecture. ~\cite{lstm_esteban} We took inspiration from their architecture and decided to deploy an LSTM based neural network architecture, shown in figure \ref{fig:lstm_arch}, as one of our learning model.

Even after knowing the fact that XGBoost is a go-to model for data science competitions, we somewhere lacked the intuition on why XGBoost works magically well in such competitions and machine learning in general. And this gap was filled by another well articulated data science blog, \textbf{\textit{'XGBoost Algorithm: Long May She Reign!'}}.~\cite{xgboost}

\section{Data}

We followed the conventional data engineering process where we started with data cleaning followed with exploratory data analysis and feature engineering. 


\subsection{Data cleaning}

First things first, we clipped the prediction feature \textit{Item Count} to match the prediction output requirement \textit{[0,20]}. Next step was to fix the invalid values and missing values. There weren't many corrupted values; the data set was quite legitimate in this sense. However, for the minor fixes that we did, our approach was to either remove it completely or to interpret the values by taking mean or median. We took the most appropriate approach on a case by case basis.

\subsection{Exploratory Data Analysis (EDA)} 

\begin{figure}[ht]
\vspace{0 mm}
    \centering
   \includegraphics[width=200pt]{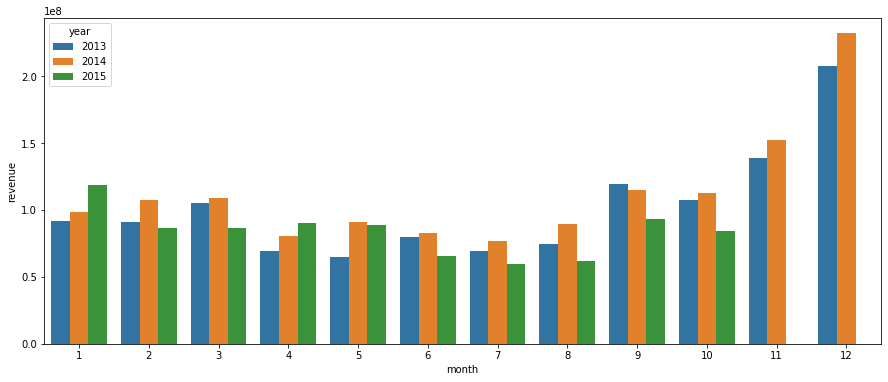}\includegraphics[width=200pt]{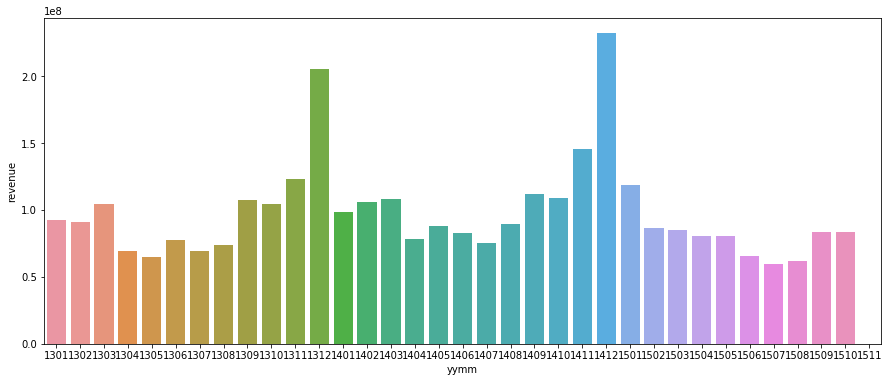}
  \caption{Plots of Revenue vs Month}
  \label{fig:plot_eda}
\end{figure}

To get a lead for the features to derive, EDA was a required step. The plots which gave us pretty good lead for data preparation and feature extraction are shown in figure \ref{fig:plot_eda}. We noticed some key observations that we applied during feature engineering process. The most important finding was that the trend is normally distributed with just few peaks in between which are observed yearly (useful for getting intuition of extracting time lag). Some other observations we noticed were that sales is highest during weekends so features that might be helpful to capture the effect might be a count of each of the weekday in the date-block, and count of weekends. We further noticed that revenue peaks towards end of the year and is lower in mid-year. Moreover, the breakdown of revenue by category varies considerably month to month.

\subsection{Feature engineering}

Taking inspiration from the briefly mentioned EDA process, the following features were generated:
\begin{enumerate}
    \item First step was to create merged data set by combining all the provided data sets i.e. shops, items and item categories data set into one master data set.
   \item Transforming the combined master data set from Time Series Format into Grid Format
   \begin{itemize}
     \item Created a grid to store data with unique shop\_id, item\_id combinations for each date\_block\_number.
     \item Since the sales data only has rows with transactions, the grid was filled with 0 for item shop combinations that are not sold for the month (date\_block\_number)
   \end{itemize}
   \item Revenue related features
   \begin{itemize}
     \item aggregated total revenue for the month by ['shop\_id', 'date\_block\_number']
     \item aggregated total revenue for the month by shop\_id + category\_id
   \end{itemize}   
   \item Item count related features
    \begin{itemize}
     \item aggregated total item\_count\_day for the month by shop\_id.
     \item aggregated total item\_count\_day for the month by shop\_id + category\_id
     \item aggregated total item sold after grouping it by ['date\_block\_number', 'item\_id'] and cumulative item sold.
     \item flag new items having 0 cumulative sale in previous month, but > 1 in current month.
   \end{itemize} 
   \item Price related features.
   \begin{itemize}
     \item Computed weighted mean price feature for the [item\_id, date\_block\_number].
     \item Computed average price feature for the [item\_id, date\_block\_number].
     \item Computed average price for the [item\_id, date\_block\_number, store\_id].
   \end{itemize} 
   \item Extracting lags and its interactions
    \begin{itemize}
     \item Lags of all the numeric features were calculated for the past 1-3 months and 12 months.
     \item To further get some more interaction between lags, we computed difference between the above extracted lags of numeric features and added that as an additional features.
   \end{itemize} 
  \item One hot encoding 'month', 'year', 'item\_category\_id', 'shop\_id' and their various combinations.
\end{enumerate}

Finally, we concluded the data engineering part by splitting the data sets into training set and validation set. where we split the first 32 months for training and the 33rd month data was sliced out for validation.





\section{Experiments}

\subsection{Models}
\subsubsection{XGBoost}
 XGBoost was one of the starting algorithms we chose. As per statistics, it is most widely used model for Kaggle competitions. It provides system optimization through parallelization and hardware optimization.
 XGBoost has advantages over general gradient boosting in terms of providing regularization through  a combination of both the ridge and lasso regressions. It also handles different type of sparsity patterns in the data efficiently. Table \ref{table:xg_params} enumerates the hyperparameters we have tuned over the validation set using uniform random sampling over a grid search space.
\begin{table}[ht]
 \centering
\begin{tabular}{|l|l|c|}
\hline
\textbf{Parameters}         & \textbf{Description}         & \textbf{Optimal Value}                                          \\ \hline
eta                & Learning Rate         & 0.148                                       \\ \hline
max\_depth         & Maximum depth of the tree         & 6                             \\ \hline
min\_child\_weight & Minimum sum of weight of observation in a child         & 26 \\ \hline
lambda             & L2 regularization term on weights  & 0.171                     \\ \hline
alpha              & L1 regularization term on weights   & 0.170                    \\ \hline
\end{tabular}
\vspace{2 mm}
\caption{Parameters of XGBoost model tuned over validation set}
\label{table:xg_params}
\end{table}



\subsubsection{ARIMA model}
ARIMA modelling is usually a go-to approach for time series data. But ARIMA only works best for univariate time series data i.e. the one which has a single variable. For e.g., stock market prediction. To apply ARIMA for this dataset, our approach was to group the time series data with respect to the identifier columns and fit ARIMA for each group and make predictions. This required us to fit the model multiple times for as many unique values in identifier we have.

\vspace{-1 mm}

\subsubsection{Long Short-Term Memory(LSTM) network}
\begin{figure}[ht]
\vspace{0 mm}
    \centering
  \includegraphics[width=1.0\columnwidth]{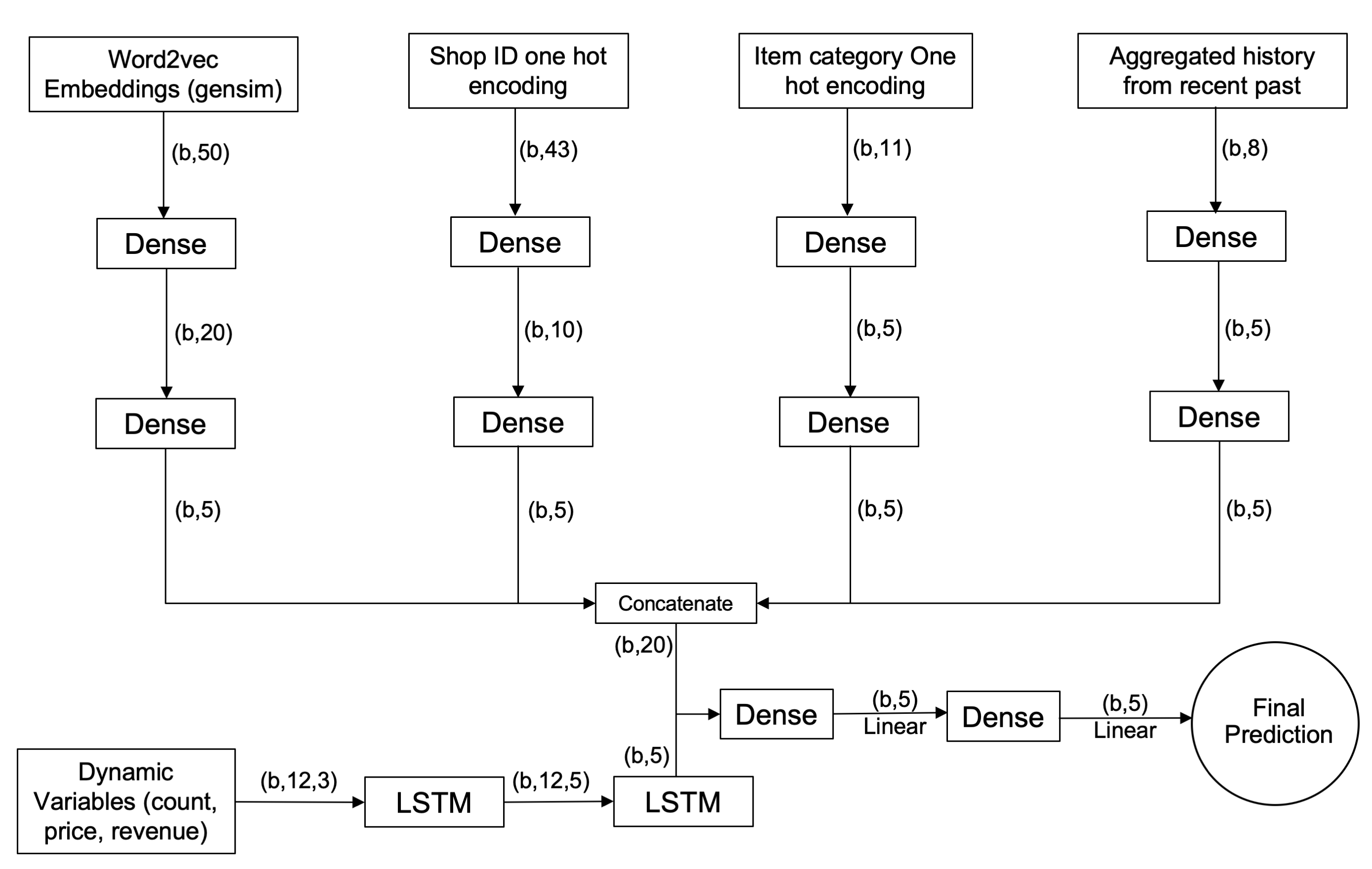}
  \caption{Customized LSTM-based neural network architecture, with batch size (b) and tanh activation}
  \label{fig:lstm_detail}
  \vspace{0 mm}
\end{figure}

Given the static and dynamic nature of input features, a customized neural network architecture (figure \ref{fig:lstm_detail}) is subsequently used for learning this regression task. The proposed architecture resulted in a decrease of space requirement by a factor of 12 over plain LSTM, where all static features are replicated across time steps. For the purpose of achieving optimal values for hyperparameters, we have created a grid search space. Because of the time and resource constraint, only batch size and L2 regularization coefficient $(\lambda)$ is considered during hyperparameter tuning. Also, it is important to note here that the value of $\lambda$ is assumed equal for all layers. The optimal value determined for batch size is 512 samples, and that for $\lambda$ is 0.001. Adam optimizer and mean squared loss is used in optimization algorithm, which is executed over 5 epochs of training data.
\vspace{-1.25 mm}
 

\subsection{Evaluation}

The evaluation metric used to evaluate the performance of different models is the Root Mean Square Error (RMSE). 
The training and validation data were split based on description in Section 3.3. For the test set, we provided the prediction for November 2015 (month 34), and the same has been used to calculate our public leader board position on Kaggle.



\subsection{Setup}

We implemented our models in Google Colab with 25 GB RAM and GPU support. We have used python as the coding language for this project. The different libraries and packages that empowered our modelling process were: numpy, pandas, sklearn, keras, tensorflow, matplotlib, and xgboost. 


\section{Results}
Our best performing model is the XGBoost, whose performance along with the other models is listed in table \ref{tab:results}. The similar values for training and validation RMSE suggests low variance across the models we tried. Table \ref{tab:feature_imp} lists relative importance of top 10 features in XGBoost algorithm based on their F-score. It can be seen that the derived features with lag have a large influence on predictions.

\begin{table}[ht]
\centering
\begin{tabular}{|l|l|l|l|}
\hline
Model   & Training RMSE & Validation RMSE & Test RMSE \\ \hline
XGBoost & 0.764467      & 0.807264        & 0.87815   \\ \hline
LSTM    & 0.804657      & 0.889786        & 0.92417   \\ \hline
ARIMA   & 0.963426      & 0.982234        & 1.09266   \\ \hline
\end{tabular}
\vspace{2 mm}
\caption{Performance of different models}
\label{tab:results}
\end{table}
\vspace{-5 mm}

\begin{table}[ht]
\centering
\begin{minipage}{.4\textwidth}
\begin{tabular}{|l|l|}
\hline
\textbf{Features}           & \textbf{F-Score} \\ \hline
item\_id                    & 4168             \\ \hline
target\_item\_lag\_1        & 3034             \\ \hline
enc\_shop\_id\_category     & 2845             \\ \hline
item\_category              & 2738             \\ \hline
date\_block\_num            & 2515             \\ \hline
\end{tabular}
\vspace{2 mm}
\end{minipage}
\begin{minipage}{.4\textwidth}
\centering
\begin{tabular}{|l|l|}
\hline
\textbf{Features}           & \textbf{F-Score} \\ \hline
target\_category\_lag\_1    & 2393             \\ \hline
enc\_shop\_id               & 2368             \\ \hline
target\_shop\_trend\_1\_2   & 2278             \\ \hline
target\_price\_mean\_lag\_1 & 2222             \\ \hline
month                       & 2194             \\ \hline
\end{tabular}
\vspace{2 mm}
\end{minipage}
\caption{Top features in XGBoost}
\label{tab:feature_imp}
\end{table}

\vspace{-1 mm}
\section{Conclusion and future work}

Sales forecasting has become highly sophisticated and plays a vital role to the competitiveness of many companies. We have tried to deploy different learning algorithms for forecasting sales in this paper. XGBoost performed best in achieving our objective on this dataset. This can be attributed to the good feature engineering and the ability to try out a large range of hyperparameters during optimization. 


We have seen that selecting optimal parameters for a neural network architecture can often make the difference between ordinary and state-of-the-art performance.\cite{lstm_hyper} Thus, a more rigorous approach for hyperparameters selections can be used to tune the customized LSTM-based architecture. Finally, we would also like to use ensemble techniques in future to combine predictions from the different models we tried; since that may potentially reduce both bias and variance in our output predictions.

\bibliographystyle{unsrtnat}
\bibliography{main}

\begin{thebibliography}{7}
\providecommand{\natexlab}[1]{#1}
\providecommand{\url}[1]{\texttt{#1}}
\expandafter\ifx\csname urlstyle\endcsname\relax
  \providecommand{\doi}[1]{doi: #1}\else
  \providecommand{\doi}{doi: \begingroup \urlstyle{rm}\Url}\fi

\bibitem[Kaggle(2018)]{kaggle}
Kaggle.
\newblock Predict future sales.
\newblock Online, 2018.
\newblock URL
  \url{https://www.kaggle.com/c/competitive-data-science-predict-future-sales/overview}.

\bibitem[Bandara et~al.(2019)Bandara, Shi, Bergmeir, Hewamalage, Tran, and
  Seaman]{lstm}
Kasun Bandara, Peibei Shi, Christoph Bergmeir, Hansika Hewamalage, Quoc Tran,
  and Brian Seaman.
\newblock Sales demand forecast in e-commerce using a long short-term memory
  neural network methodology.
\newblock arXiv:1901.04028, 2019.

\bibitem[Kihoro et~al.(2004)Kihoro, Otieno, and Wafula]{arima}
J.M. Kihoro, R.O. Otieno, and C.~Wafula.
\newblock Seasonal time series forecasting: A comparative study of arima and
  ann models.
\newblock African Journal of Science and Technology (AJST) Science and
  Engineering Series Vol. 5, No. 2, pages: 41-49, 2004.

\bibitem[Eno5(2018)]{kaggle_blog}
Eno5.
\newblock Time series - arima, dnn, xgboost comparison.
\newblock Online, 2018.
\newblock URL \url{https://bit.ly/3fvCPtk}.

\bibitem[Esteban et~al.(2016)Esteban, Staeck, Yang, and Tresp]{lstm_esteban}
Cristóbal Esteban, Oliver Staeck, Yinchong Yang, and Volker Tresp.
\newblock Predicting clinical events by combining static and dynamic
  information using recurrent neural networks.
\newblock arXiv:1602.02685, 2016.

\bibitem[Morde(2019)]{xgboost}
Vishal Morde.
\newblock Xgboost algorithm.
\newblock Online, 2019.
\newblock URL \url{https://bit.ly/3fCSZ4h}.

\bibitem[Reimers and Gurevych(2017)]{lstm_hyper}
Nils Reimers and Iryna Gurevych.
\newblock Optimal hyperparameters for deep lstm-networks for sequence labeling
  tasks.
\newblock arXiv:1707.06799, 2017.

\end{thebibliography}

\end{document}